\documentclass{article}

\usepackage{theorem}

\usepackage{arxiv}
\usepackage{upgreek}
\usepackage[utf8]{inputenc} % allow utf-8 input
\usepackage[T1]{fontenc}    % use 8-bit T1 fonts
\usepackage{hyperref}       % hyperlinks
\usepackage{url}            % simple URL typesetting
\usepackage{booktabs}       % professional-quality tables
\usepackage{amsfonts,mathabx}       % blackboard math symbols
\usepackage{nicefrac}       % compact symbols for 1/2, etc.
\usepackage{microtype}      % microtypography
\usepackage{lipsum}		% Can be removed after putting your text content
\usepackage{graphicx}
\usepackage{natbib}
\usepackage{doi}
\usepackage[ruled,linesnumbered]{algorithm2e}

\newtheorem{theorem}{Theorem}

\newtheorem{definition}[theorem]{Definition}

\title{A Multi-objective Evolutionary Algorithm for  \\ EEG Inverse Problem}

\author{ \href{https://orcid.org/0000-0002-8313-6338}{\includegraphics[scale=0.06]{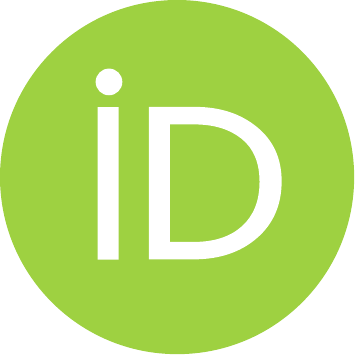}\hspace{1mm}Jos\'e Enrique Alvarez-Iglesias}\thanks{Corresponding author.} \\
	Department of Neurophysics\\
	Cuban Center for Neuroscience\\
	\texttt{jose.alvarez@cneuro.edu.cu} \\
	\And
	\hspace{1mm}Mayrim Vega-Hernandez \\
	Department of Neurophysics\\
	Cuban Center for Neuroscience\\
	\texttt{mayrim@cneuro.edu.cu} \\
	 \AND
	 \hspace{1mm}Eduardo Mart\'inez-Montes \\
	 Department of Neurophysics\\
	 Cuban Center for Neuroscience\\
	 \texttt{eduardo@cneuro.edu.cu} \\
}

\renewcommand{\shorttitle}{\textit{arXiv} Template}

\hypersetup{
pdftitle={A template for the arxiv style},
pdfsubject={q-bio.NC, q-bio.QM},
pdfauthor={David S.~Hippocampus, Elias D.~Striatum},
pdfkeywords={evolutionary algorithm, multiobjective, inverse problem, EEG},
}

\begin{document}
\maketitle

\begin{abstract}
	In this paper, we proposed a multi-objective approach for the EEG Inverse Problem. This formulation does not need unknown parameters that involve empirical procedures. Due to the combinatorial characteristics of the problem, this alternative included evolutionary strategies to resolve it. The result was a Multi-objective Evolutionary Algorithm based on Anatomical Restrictions (MOEAAR) to estimate distributed solutions. The comparative tests were between this approach and 3 classic methods of regularization: LASSO, Ridge-L and ENET-L. In the experimental phase, regression models were selected to obtain sparse and distributed solutions. The analysis involved simulated data with different signal-to-noise ratio (SNR). The indicators for quality control were Localization Error, Spatial Resolution and Visibility normalized. The MOEAAR evidenced better stability than the classic methods in the reconstruction and localization of the maximum activation. The norm L0 was used to estimate sparse solutions with the evolutionary approach and its results were relevant.
\end{abstract}

% keywords can be removed

\section{Introduction}
Electroencephalogram (EEG) is a neuroimage technique which register the electromagnetic field created on the cerebral cortex. The sources are the synchronized post-synaptic potentials between interconnected neurons. The activations are caused by large micro-scale masses of neuronal activity. However, they are small generating sources compared to the range obtained by an electrode during its recording \cite{Sri19}. In the EEG, the sensors are organized like a discrete array on the scalp. This is a non-invasive method with high temporal resolution. Although, the measurements are distorted by tissues of the head that define a non-homogeneous conductive volume. This condition introduce difficulties in the localization process of generated sources. The EEG Inverse Problem (IP) proposes to estimate the origin of the measurements recorded on the scalp \cite{Sri19}.

The solution of EEG IP needs a forward model that represents the electromagnetic field created by neuronal activations and their effects on the scalp. The best-known approaches approximate large neural masses like a finite set of points (dipoles). The relationship between these dipoles and the electrodes is linear when the model is discretized by Maxwell equations. The outcome is a linear regression \cite{Paz17}:

\begin{equation}
	V^{m\times n}=K^{m\times n}J^{n\times t}+ e^{m\times t}
	\label{eq:1}
\end{equation}

where $V$ is a matrix of $m\times t$ with the potential differences between $m$ electrodes and the sensor of reference for each instant of time ($t$). The matrix $K^{m\times n}$ represents the relationship between $n$ possible sources and the recorded values obtained by $m$ electrodes. The columns of $K$ are calculated with the conductivity and the geometry of intermediate layers between the cerebral cortex and the scalp. For example: the skin, the skull and the cerebrospinal fluid\cite{Nu06}. The matrix $J^{n\times t}$ represents a Primary Current Density Distribution (PCD) of $n$ dipoles in $t$ instants of time. 

In general, the number of sensors is too small (in the order of tens) with respect to the amount of possible sources (in the order of thousands). For this reason, the system (\ref{eq:1}) is indeterminate and an ill-posed problem\cite{Veg19}. In the matrix $K$, the columns associated to close sources produce similar coefficients. These high correlations cause the bad condition of the system and increase their sensibility on noisy data. In nature problems, the noise is an inherent factor and the EEG signal is not an exception. Thus, this research proposed evolutionary strategies with more tolerance to noisy data than numerical methods based on mathematical models \cite{Qian15}.

Electromagnetic Source Imaging (ESI) represents the approaches proposed to resolve the EEG IP. Due to the bad condition of the linear equation, their include a priori information in the optimization process. Regularization models minimize the linear combination between the squared fit and penality functions. These constraints represent assumptions on the truth nature of the sources. A classic model to obtain sparse estimations is LASSO (Least Absolute Shrinkage Selection Operator) which defines the norm of L1 space as a penalty function $L1$ ($\|J\|_1$):

\begin{equation}
	\min_J \|V-KJ\|^2_2 + \lambda\|J\|_1
	\label{eq:2}
\end{equation}

However, L0 ($\|\cdot\|_0: \mathbb{R}^{n\times 1}\rightarrow \mathbb{N}$) is the right norm to model sparsity and L1 is just an approximation. In \cite{Nata15}, B.K. Natarajan proved how the complexity of (2) with the norm L0 as a penality function is NP-Hard in Combinatorial Optimization. Unfortunately, this penality function is discontinuous and non-convex in $\mathbb{R}$.

The regularization models allow to combine the objective function with other restrictions to obtain smoothness and nonnegative coefficients. In the field, some relevant methods are: ENET (Elastic Net) \cite{Zou05}, Fused LASSO (fussed LASSO) \cite{Tib05}, Fusion LASSO (LASSO fusion), SLASSO (Smooth LASSO) \cite{Land96} and NNG (Nonnegative Garrote) \cite{Gij15}. They are particular derivations from the expression \cite{Veg19}:

\begin{equation}
\hat{J}=argmin_{J} \{(V-KJ)^T(V-KJ)+\Uppsi(J)\},
\label{eq:3}
\end{equation}

where $\Uppsi(J)=\sum_{r=1}^{R} \lambda_r \sum^{N_r}_{i=1} g^{(r)}(|\theta^{(r)}_i|)$ and $\theta^{(r)}=L^{(r)}J$. A more compact equation is:

\begin{equation}
	\min_{J}\, \|V-KJ\|^2_2 + \sum_{r=1}^{R} \lambda_rp^{(r)}(\theta^{(r)})
	\label{eq:4}
\end{equation}

where the objective function is a weighted sum of the squared error and the penalty functions $p^{(r)}(\theta^{(r)})=\sum_{i=1}^{N_r}g^{(r)}(|\theta^{(r)}_i|)$ such that $r\in \{1,\dots,R\}$. The $L$ matrix is structural information incorporated into the model and represents possible relationships between variables. For example: Ridge L is a derivative model with the norm L2 ($p$) and a Laplacian matrix ($L$) of second derivatives. The regularization parameter $\lambda$ is unknown and its estimation involve heuristic methods to choose the optimal value\cite{Paz17}. For example, the optimal solution for the Generalized Cross-Validation (GCV) function\cite{Veg19}. Although, this method does not guarantee the success for all cases, it proposes solutions good enough to the problem. Multi-objective approaches discard these limitations and explore the solutions space guiding by the best compromise between objective functions. The next section introduces relevant concepts in the multi-objective theory. The section 3 describes the proposed evolutionary approach and dedicates a brief subsection to each stage.

\section{Muti-objective optimization}
The Multi-objective optimization estimates a set of feasible solutions that conserve high evaluations and equal commitment among the objective functions. These functions have to share the same priority in the model and the solution methods have to avoid preferences in the search process. The multi-objective version for regularization models incorporates the same functions to a vector of objectives. The unrestricted bi-objective model for LASSO is:

\begin{equation}
	\min (\|V-KJ\|^2_2, \|J\|_1)
	\label{eq:5}
\end{equation}

Multi-objective formulations do not use unknown auxiliary parameters like regularization models. This advantage allows them to solve problems with poor information about the relationship among the objective functions. Some approaches propose the model \ref{eq:5} to solve inverse problems in image and signal processing \cite{Gong16, Gong17}. The multi-objective model for the equation \ref{eq:4} without regularization parameters is:

\begin{equation}
	\min_J (\|V-KJ\|^2_2, p^{(1)}(\theta^1),\dots, p^{(R)}(\theta^R)) \qquad R\in\mathbb{N}
	\label{eq:6}
\end{equation}

Thus, a multi-objective version for each regularization model is possible to obtain changing the functions $p^{(i)} (\theta^i)$. For example: the equation \ref{eq:5} assumes that $R=1$, $g^{(1)}(x)=x$ and $L^1= I_n$; therefore $p^{(1)} (\theta^1)= \|J\|_1$.

In multi-objective optimization, the definition of Pareto Dominance establishes a set of optimal solutions (Pareto Optimal Set) for the model. If $F:\Omega\rightarrow \Lambda$ is a vector of functions, $\Omega$ is the space of decision variables and $\Lambda$ is the space defined by the objective functions, then \cite{Coe07}:

\begin{definition}[Pareto Dominance]
	\label{def:1}
	A vector $u=(u_1,\dots,u_k)$ dominates another vector  $v=(v_1,\dots,v_k)$ (denoted as $u\preceq v$) if and only if $u$ is partially less than $v$, that is, $\forall i \in \{1,\dots,k\}$, $u_i\leq v_i \wedge \exists j\in \{1,\dots,k\}:u_j<v_j$.
\end{definition}

A solution for a Multi-Objective Problem (MOP) is defined as:

\begin{definition}[Pareto Optimality]
	\label{def:2}
	A solution $x\in \Omega$ is Pareto Optimal on the $\Omega$ space if and only if $\nexists x'\in \Omega$ for which $v=F(x')=(f_1(x'),\dots,f_k(x'))$ dominates $u=F(x)=(f_1(x),\dots,f_k(x))$.
\end{definition}

A set of the solutions is defined as:

\begin{definition}[Pareto Optimal Set]
	\label{def:3}
	For a given MOP with $F(x)$ vector of objective functions, the Pareto Optimal Set $P^{*}$, is defined as:
	\begin{equation}
	P^{*}:=\{x\in\Omega\, |\, \nexists\, x'\in \Omega ,\; F(x')\preceq F(x)\}
	\label{eq6}
	\end{equation}
\end{definition}

Then, $F(P^*)\subset \Lambda$ defines a Pareto Front in the image space. The formal definition is:

\begin{definition}[Pareto Front]
	\label{def:6}
	For a MOP, the vector of objectives functions $F(x)$ and the Pareto Optimal Set $P^*$, the Pareto Front $FP^*$ is defined as:
		\begin{equation}
		FP^*:=\{u=F(x)\,|\,x\in P^*\}
		\label{eq7}
	\end{equation}
\end{definition}

The proposed algorithm searches the $P^*$ in the solution space that guarantees a $PF^*$ with uniform distribution between its points. For MOPs that require only one solution, the researchers have proposed to apply a selection criterion to choose a result from the $PF^*$. Desicion Maker (DM) is the procedure to select a solution from $PF^*$ and some algorithms used at the beginning (a priori), at the end (a posteriori) or during the execution (progressive) \cite{Coe07}.

\section{Method}

Evolutionary algorithms estimate solutions to  multi-objective problems with high quality in nature life \cite{Coe07}. These strategies allow the hybridization with other algorithms to exploit heuristics information on the problem. In this research, the proposed Multi-objective Evolutionary Algorithm (MOEA)  combines co-evolutionary strategies and local search techniques to resolve the EEG IP. In runtime, the algorithm holds a set of solutions that updates in each iteration. The evaluation over the objective functions and the theory of the problem establish the comparison between solutions. However, multi-objective models have to use a comparison criterion based on Pareto Dominance. The stability of the algorithm was evaluated in different levels of noise over simulated data. The quality measurements were Localization Error, Spatial Resolution and Visibility. 

\subsection{Evolutionary Multi-objective Algorithm based on Anatomical Restrictions}

\begin{figure}
	\centering
	\includegraphics[scale=0.6]{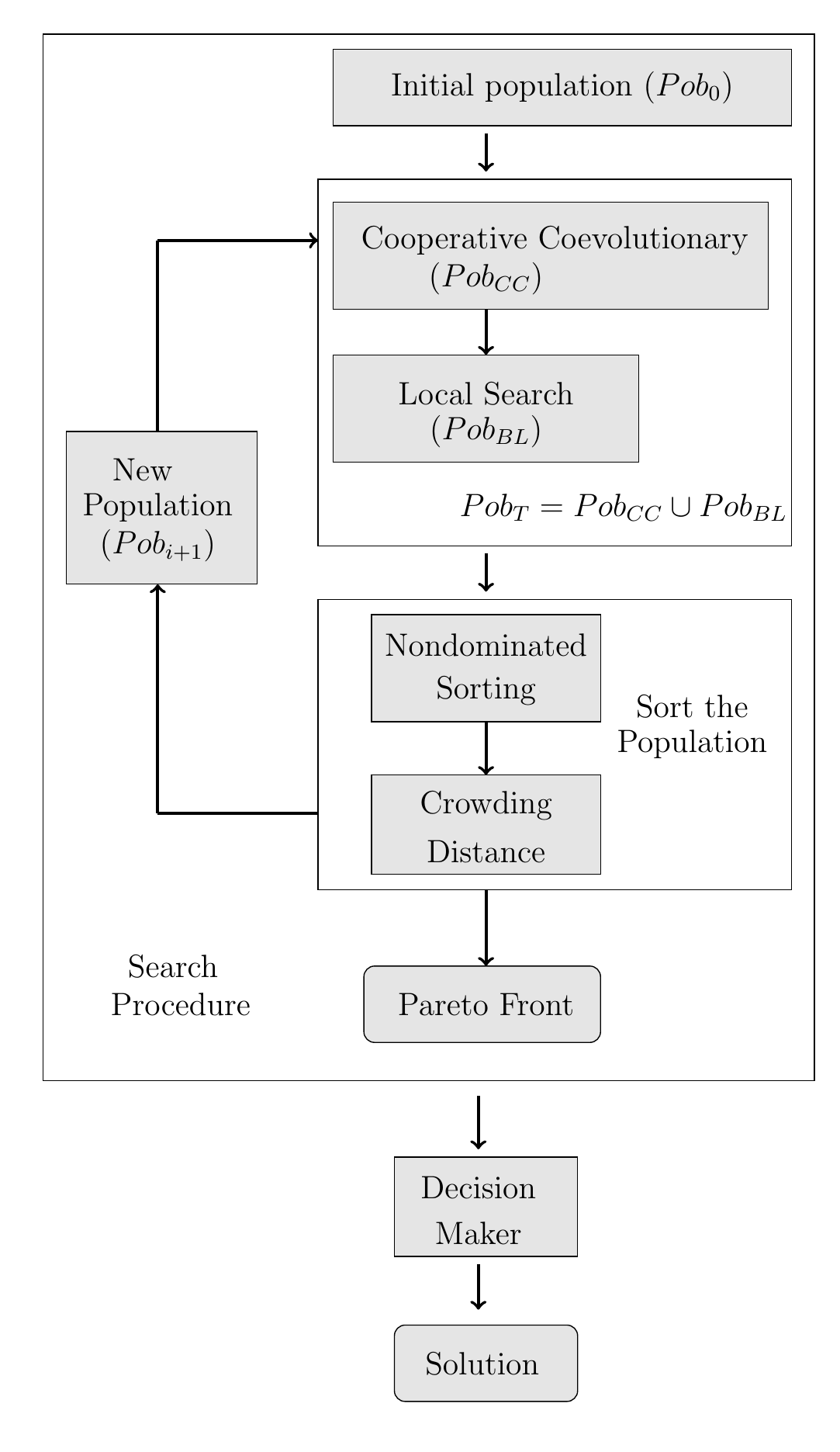}
	\caption{Diagram of MOEAAR with general steps. The workflow is divided in 2 general stage: Search Procedure and Decision Maker. The first one explores the solution space to find relevant solutions that conserve a commitment among the objectives. The second one selects the result from the last $FP^*$ obtained in the previous stage.}
	\label{fig:1}
\end{figure}

This evolutionary approach incorporated a coevolutionary strategy based on anatomical constraints to generate a new set of solutions. For this reason, the method was named Multi-objective Evolutionary Algorithm based on Anatomical Restrictions (MOEAAR). In addition, the procedure included a local search method to improve accuracy and intensify the search in specific areas of the solution space. The last stage applied a Decision Maker which select the result from a set of possible candidates ($P^*$). The Figure \ref{fig:1} customizes the stages of the algorithm in a diagram. 

\subsection*{Initial Population}
The Search Procedure maintains a set of solutions with improvements obtained between iterations. The solutions are represented as individuals in a population ($Pob_i$). The initial population ($Pob_0$) is made up of N individuals representing each region of interest (ROI). An individual represents the region that limits the sources with activations in the associated solution. The value of the activations to create the initial population was the norm L1 of the pseudoinverse ($K^+$) solution \cite{Dem97}.

\subsection*{Cooperative Coevolutionary}
The Cooperative Coevolutionary methodology divided a high-dimensional problem into subproblems with fewer decision variables. In the Coevolution stage, these subsets of variables defined new subpopulations by projecting the original individuals to the subspaces associated. Then, the Search Procedure began with an evolutionary algorithm which generates new candidates with just one step by subproblem. In this approach, NSGA II was selected for the task. The result was the combination of successive improvements through independent optimizations by subpopulation. For this reason, a Cooperation process was applied to estimate relevant solutions in the evaluation stage between iterations. This step required a context vector (CV) with information on features improved in previous iterations.

This strategy emphasizes the search based on correlated variables that share similar functions. In the EEG IP, anatomical constraints allow clustering possible generating sources. The algorithm proposed follow the same criterion. In the Algorithm \ref{algo:1}, a pseudocode customizes the steps of the Cooperative Coevolutionary procedure. In the line 2, a partition is established by an anatomical atlas from the Montreal Neurological Institute. In the line 5, a subset of variables ($G_j$) define a subpopulation ($pop_i^j$) by projection. Then, NSGA II (EA function) generate news individuals with highest quality from $pop_i$. Afterwards, the variables of $G_j$ are updated in VC with $best_j$. The steps 4-9 are repeated $t$ iterations until stop.

\begin{algorithm}
	\DontPrintSemicolon
	\KwData{$pop_i$(last population), $f$(objective function)}
	\KwResult{$pop_{new}$}
	\Begin
	{
		$P(G)=\{G_1,\dots,G_k\} \leftarrow \textnormal{partition(n)}$ \;	 	 	
		\While{condition()}
		{
			\ForEach{$G_j \in P(G)$}
			{
				$pop_i^j\leftarrow \textnormal{subPopulation}(pop_i,G_j)$\;
				$[best_j, popR]\leftarrow \textnormal{EA}(f,pop_i^j,VC,G_j)$\;
				$pop_{new}\leftarrow \textnormal{sortPopulation}(popR,G_j)$\;
				$VC \leftarrow \textnormal{updateVC}(best_j,G_j)$\;			
			}
			$pop_i\leftarrow pop_{new}$\;
		}	 	
	}
	\caption{CC($pop_i,f$)}
	\label{algo:1}
\end{algorithm}

The NSGA II used reproduction operators to generate new individuals (or solutions) between iterations. Reproduction operators are designed to ensure the exploitation and exploration of promising regions in a feasible solutions space \cite{Coe07}. First, a Binary Tournament procedure with repetitions selected the individuals to generate new solutions. Then, an arithmetic crossover operator generated new individuals from the selection result. A new individual is defined as:

\begin{equation}
d_i=\alpha p^1_i + (1-\alpha)p^2_i.
\end{equation}

where $d=\{d_1,d_2,\dots,d_n \}$  is the solution vector for the newest individuals and $p^k=\{p_1^k,p_2^k,\dots,p_n^k\}$ are the selected parents such that $k\in \{1,2\}$. The values of $\alpha$ were generated randomly following a normal distribution $N(0,1)$.

A mutation procedure was the next step to guarantee more diversity in the evolution process. The option selected was the polynomial mutation and new individuals were obtained like:

\begin{equation}
d_i^{(t+1)}=d_i^{(t)} + \sigma^{(t+1)}\cdot N(0,1).
\end{equation}
	
The perturbation $\upsigma$ is the mutation step and its initial value decreases during the execution of the algorithm. This parameter controls the diversity between parents ($d_i^{(t)}$) and the new individuals ($d_i^{(t+1)}$). The mutation process obtained a new population from a subset of solutions generated by crossover operator in the previous step.

The new subpopulations are the join between individuals from crossover and mutation operator. The result was a population ($Pop_{CC}$) created by a reconstruction process that combine the subpopulations. The next stage was designed to improve the accuracy by forcing a deep search in promising areas of the solution space. For this reason, a Local Search procedure was selected for this task and a brief description can be found in the section below.

\subsection*{Local Search}

Local search (LS) algorithms are optimization techniques that search optimal solutions using neighborhood information. Among the most relevant methods for continuous problems are: the Coordinate Descent method, the Maximum Descent method and the Newton-Raphson variants. In addition, new strategies have proposed in the last decades that combine thresholding techniques with directional descent methods.

In MOEAAR, the LS was used to achieve high precision in the solutions obtained by Cooperative Coevolutive stage. This module performs a descending search on each individual in the population $Pop_{CC}$ to generate $Pop_{LS}$. The implemented method was proposed in \cite{Li14} as Local Smooth Threshold Search (LSTS). However, some adaptations were included for the EEG IP.
The method LSTS is a variant of the Smoothed Threshold Iterative Algorithm designed to obtain sparse estimations in minimization problems with L1 norm. The algorithm generated a sequence of optimal solutions $\{J^k,k=0,1,2,\dots,n\}$ where $J^{k+1}$ was obtained from $J^k$ to resolve the optimization problem:

\begin{equation}
J^{k+1}=\min_J \frac{1}{2} \|J-v^k\|^2_2 + \frac{\lambda}{\beta^k} p(J)
\end{equation}

where the diagonal matrix $\beta^kI$ is an estimation of the Hessian matrix $\nabla^2f(J^k)$ and

\begin{equation}
v^k=J^k-\frac{1}{\beta^k}\nabla f(J^k).
\end{equation}

Then, if $p(J)=\|J\|_1$:

\begin{equation}
J_i^{k+1}=soft(v_i^k,\frac{\lambda}{\beta^k})
\end{equation}

with $soft(u,a)=sgn(u)\cdot\max\{|u|-a,0\}$ is the smoothed threshold function. For the value $\lambda$ that equals both terms $(\|V-KJ\|_2^2,\|J\|_1)$ was proposed:

\begin{equation}
\hat{\lambda}=\frac{\|V-KJ\|^2_2}{\|J\|_1}.
\end{equation}

In this case, $J=J_{CC}$ was considered the best-fit estimation obtained in $Pop_{CC}$. The Barizilai-Borwein equation was proposed to obtain the parameter $\beta^k$ in \cite{Li14}.

The result was a new population $Pop_{LS}$ and a temporary set defined as $Pop_T=Pop_{CC} \cup Pop_{LS}$ such that $num(Pop_T)=2\cdot N$. However, this set is twice huger than the initial population ($N$). Thus, the next step was to reduce $Pop_T$ to $N$ individuals with the highest evaluation and compromise between the objective functions. First, the individuals were ordered by the operators: Non-dominated Sorting \cite{Zha01} and Crowding Distance \cite{Deb02, Chu10}. Then, the $N$ individuals more relevant were selected and the rest rejected to conform a new population with the best from $Pop_T$. Also, this process generated a Pareto Front ($PF_{know}$) that was updated in each cycle. A Decision Maker process selected the result of the algorithm from the last $PF_{known}$.

\subsection*{Desicion Maker}
The Decision Maker process select the solution with more compromise from Pareto Front. In this approach, a first stage selected the region with more instances among the solutions of the $PF_{know}$. This criterion of convergence between structures is justified by the trend towards a uniform population in the last iterations. Then, the Pareto Optimal Set was reduced to solutions that converge to the selected structure. In other words, the $ROI^*=ROI_{i}$ such that $i= arg\max_i cantRep(ROI_i)$ where:

\begin{equation}
cantRep(ROI_i)=\sum_{i=1}^{|FP_{know}|} g(ROI_i,Ind_i)
\end{equation}

\begin{equation}
g(ROI,Ind)=
\left\lbrace
\begin{array}{ll}
1,& \exists v\in\mathbb{R}^3, [v\in ROI]\wedge\\
&\hspace{30pt} [Ind(C^{-1}(v))\in\mathbb{R}^{*}] \\
&\\
0,& \textnormal{otherwise}.
\end{array}	
\right.
\end{equation}

where that $Ind$ is an individual of the Pareto Optimal Set associated with $PF_{know}$, $C:I\leftarrow R^3$ is a mapping function and $I=\{1,\dots,n\}$ is the set of indexes. The result was obtained from the elbow on the B-Spline curve that interpolates the selected points from the filter process on the $PF_{know}$.

\section{Results}

The study was designed to compare MOEAAR with 3 classic methods selected for their relevance in the resolution of EEG Inverse Problem. The testing data included 16 simulations with different topological characteristics. In the tests, MOEAAR performed 100 iterations in the Search Procedure before estimate the PF. In the coevolutionary stage, the crossover operator generated new individuals with the $80\%$ of the individuals selected with repetition. However, the mutation process just involved the $50\%$. The comparative analysis was among MOEAAR solutions and the estimations proposed by Ridge-L, LASSO and ENET-L. The evolutionary approach was tested with different models (L0, L1 and L2) and each combination defined a method.

The simulations were superficial with different amplitudes bounded by 5 $nA/cm^2$ and extensions lower than two millimeters. The approximated surface (cortex) was a grid with 6,000 vertices and 12,000 triangles. However, the model just accepted 5656 possible generators and the matrix $K$ was calculated using the mathematical model of three-spheres concentric, homogeneous and isotropic\cite{Han07}. In addition, the sensor distribution was a 10-20 assembly with 128 electrodes. The sources were located in 4 regions or structures: Frontal (Upper Left), Temporal (Lower Left), Occipital (Upper Left) and Precentral (Left). Sources can be classifying in two types by extension: Punctual or Gaussian. The first one does not involve others activations. However, the second one follow a similar behavior to Normal distribution creating activations in voxels around the source. For each region, two representative simulations were created with Punctual and Gaussian extension for the same generator. Then, the equation \ref{eq:1} was applied to generate two synthetic data $V$ (SNR=0 and SNR=3) for each vector $J$. The result was a testing set which involve 16 synthetic data with differences in three factors: region, extension type and level of noise. The indicators for evaluation were the classic quality measures in the field: Localization Error, Spatial Resolution\cite{Tru04} and Visibility   \cite{Veg08}. 

The Figure \ref{fig:2} illustrates the Localization error obtained from estimations generated by the methods. This measurement quantifies how well an algorithm estimate the right coordinates of a source on the cortex. The subgraph A compares the results on data with punctual sources for different levels of noise. In this context, the coefficients are superior to 0.9 for SNR=0 but the quality change between the classic methods for SNR=3. This condition affects the classic methods (see Ridge-L$^*$, LASSO$^*$ and ENET-L$^*$) decreasing their accuracy in some regions. However, the MOEAAR variants hold their coefficients stable and over 0.9. The same behavior shows the subgraph B for data with gaussian sources.

The Figure \ref{fig:3} shows comparative information on Visibility indicator for estimations proposed by the methods. The subgraph A characterizes the case study with a punctual source by simulation. The result highlights the estimations of LASSO with an exception in the Temporal region. Unfortunately, its quality decrease as the rest of classic methods on data with noise (SNR=3) in all regions. In this subgraph, the MOEAAR-L0 maintains relevant results on data without noise and with SNR=3. The subgraph B represents the same evaluation process for data with a gaussian source in both levels of noise. In this case, the quality between regions is different in each method and the noise affect the coefficients in all cases too. 

\begin{figure}
	\centering
	\includegraphics[scale=0.7]{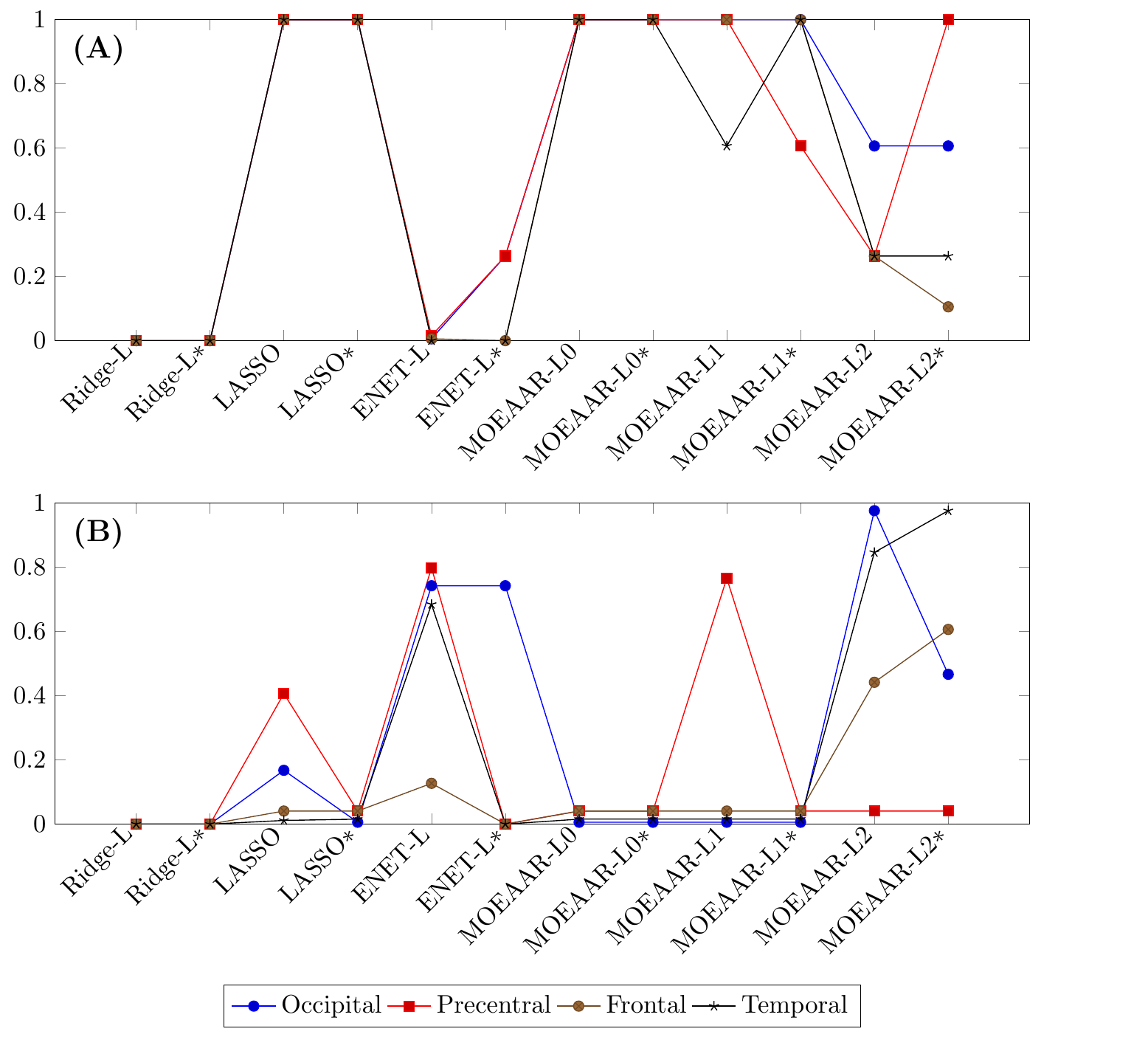}
	\caption{Spatial Resolution for Ridge-L, LASSO, ENET-L, MOEAAR-L0, MOEAAR-L1 and MOEAAR-L2. The results from data with noise (SNR=3) is represented by the name of the method with an asterisk($*$). Each line represents a simulation associated with a region. A) Results on punctual sources B) Results on gaussian sources}
	\label{fig:4}	
\end{figure}

The Spatial Resolution (SR) indicator quantify the performance of the methods to estimate the right extensions of the sources. The Figure \ref{fig:4} customized the quality of each method according to their estimations. The subgraph A shows the result from data with a punctual source and the most relevant coefficients belongs to LASSO and MOEAAR-L0. Both methods do not change too much between data with SNR=0 and SNR=3. In addition, MOEAAR-L1 proposed solutions with extension according to the punctual simulations with an exception in Temporal region for data without noise. The same difficulty decreases its quality over noised data (see MOEAAR-L1$^*$, Figure \ref{fig:4}, subgraph A) in the Precentral region. The subgraph B represents the SR data associated to the proposed estimations  for simulations with a gaussian source. The ENET-L estimate relevant solutions with coefficients over 0.6 but found some difficulty in the Frontal region. In addition, the estimations fail for data with noise (SNR=3) in Occipital, Precentral and Frontal regions. In the graph, MOEAAR-L2$^*$ proposes higher coefficients for Occipital, Temporal and Occipital regions. However, its quality decrease from 0.97647 to 0.46678 on data with noise on Occipital region.

%Both methods proved high tolerance to the noise on data with few changes in the quality.

\section{Discussion}

\begin{figure}
	\centering
	\includegraphics[scale=0.7]{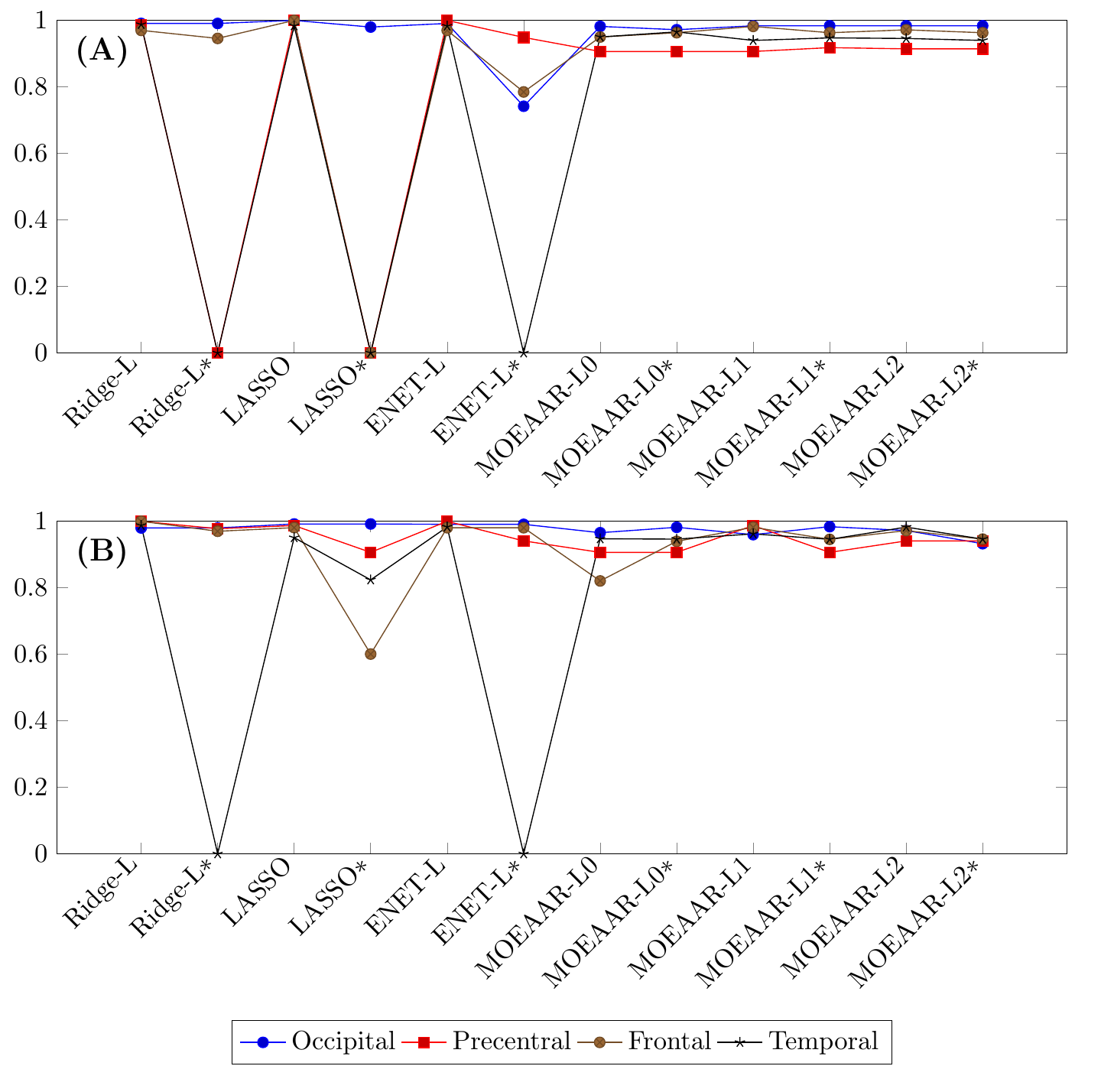}
	\caption{Localization Error for Ridge-L, LASSO, ENET-L, MOEAAR-L0, MOEAAR-L1 and MOEAAR-L2. The results from data with noise (SNR=3) is represented by the name of the method with an asterisk ($*$). Each line represents a simulation  associated with a region. A) Results on punctual sources B) Results on gaussian sources}	
	\label{fig:2}
\end{figure}

This section analyses some highlight results that stem from the previous comparative study which evidences some facts on the tolerance of MOEAAR variants to noisy data. The evolutionary approach guaranteed high accuracy and stability in the source localization (Figure \ref{fig:2}). The method maintains coefficients over 0.9 for both kind of sources: puntual and gaussian. Even, the proposed approach did not change the localization error too much with different levels of noise (SNR=0 and SNR=3). The main reason is the high tolerance of evolutionary strategies to ghost sources and noisy data. However, the classic methods worst their quality with the increasing of noise. 

\begin{figure}
	\centering
	\includegraphics[scale=0.7]{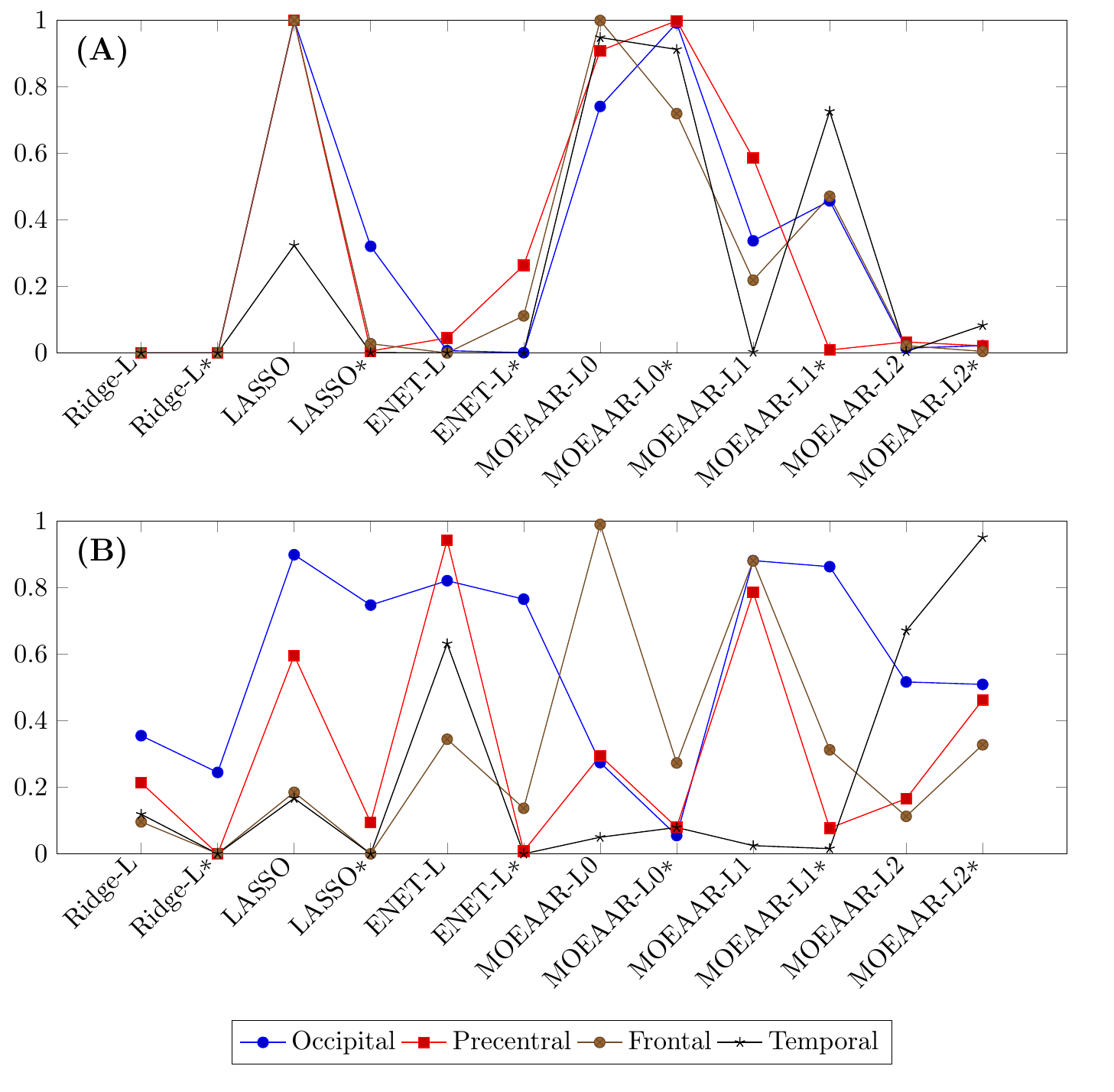}
	\caption{Visibility for Ridge-L, LASSO, ENET-L, MOEAAR-L0, MOEAAR-L1 and MOEAAR-L2. The results from data with noise (SNR=3) is represented by the name of the method with an asterisk($*$). Each line represents a simulation associated with a region. A) Results on punctual sources B) Results on gaussian sources}
	\label{fig:3}	
\end{figure}

The Visibility is a complicated indicator and the improvements in the field have been poor in the last decades. However, the result obtained in the Figure \ref{fig:3} for punctual simulations without noise highlight the quality of MOEAAR with L0. In general, this norm does not useful with the classic methods. In fact, previous researches propose approximation like SCAD, Thresholding methods or the norm L1. These efforts are justified because the right function for sparsity is the norm L0. Thus, this is the first approach that include the norm L0 in the estimation of solutions for the EEG Inverse Problem. A comparative analysis for puntual sources between MOEAAR-L0$^*$ and LASSO$^*$ in the subgraph \ref{fig:3}.A shows highest stability in MOEAAR with L0 than LASSO on noisy data. 

The Spatial Resolution indicator evaluate the capacity of the method to estimate the extension of the real source. For punctual simulations, LASSO obtained the foremost quality among the classic methods (see Figure \ref{fig:4}.A). In general, sparse methods are well-known to propose estimations with extension very close to the truth\cite{Veg19}. In addition, MOEAAR with norm L0 reaches relevant coefficients with both levels of noise (SNR=0 and SNR=3) follows by MOEAAR-L1. However, this method had some difficulties to estimate the right extension in Temporal region. In gaussian simulations, the best results were obtained between ENET-L and MOEAAR with L2 on the data without noise. Although, the last one proposed estimations with more quality for this measure than ENET-L in Occipital, Temporal and Frontal region.

\section{Conclusions}

The multi-objective formulation with a priori information allows the application of MOEAs to resolve the EEG IP without regularization parameters. The results in the Localization Error proved its capacity to find cortical and deep sources (Temporal and Precentral region). MOEAAR is an algorithm with a high stability and tolerance to noisy data in the source localization. The analysis gave reasons enough to consider the evolutionary approach as a viable method to resolve the EEG Inverse Problem. Therefore, MOEAAR could be testing with real data and Evocated Potentials in future works. The evolutionary strategies propose a way to obtain sparse solution with the norm L0.

\bibliographystyle{unsrtnat}
\bibliography{paper}  %%% Uncomment this line and comment out the ``thebibliography'' section below to use the external .bib file (using bibtex) .

\end{document}